\documentclass[10pt,twocolumn,letterpaper]{article}

\usepackage{cvpr}
\usepackage{times}
\usepackage{epsfig}
\usepackage{graphicx}
\usepackage{amsmath}
\usepackage{amssymb}
\usepackage{listings}

\usepackage[breaklinks=true,bookmarks=false]{hyperref}

\cvprfinalcopy 

\def\cvprPaperID{****} 

\begin{document}

\title{An Empirical Analysis of Approximation Algorithms for the Euclidean Traveling Salesman Problem
}

\author{Yihui He\\
Xi'an Jiaotong University\\
Xi'an, China\\
{\tt\small heyihui@stu.xjtu.edu.cn}
\and
Ming Xiang\\
Xi'an Jiaotong University\\
Xi'an China\\
{\tt\small mxiang@mail.xjtu.edu.cn}
}

\maketitle

\begin{abstract}
With applications to many disciplines, the traveling salesman problem (TSP) is a classical computer science optimization problem with applications to industrial engineering, theoretical computer science, bioinformatics, and several other disciplines \cite{introduction}. In recent years, 
there have been a plethora of novel approaches for approximate solutions ranging from simplistic greedy to cooperative distributed algorithms derived from artificial intelligence. 
In this paper,
we perform an evaluation and analysis of cornerstone algorithms for the Euclidean TSP. We evaluate greedy, 2-opt, and genetic algorithms. We use several datasets as input for the algorithms including a small dataset, a medium-sized dataset representing
cities in the United States, and a synthetic dataset consisting of 200 cities to test algorithm scalability. We discover that
the greedy and 2-opt algorithms efficiently calculate solutions for smaller datasets. Genetic algorithm has the best
performance for optimality for medium
to large datasets, but generally have longer runtime. 
Our implementations is public available \footnote{https://github.com/yihui-he/TSP}.
\end{abstract}

\section{Introduction}
Known to be NP-hard, the traveling salesman problem (TSP) was first formulated in 1930 and is one of the most studied optimization problems to date \cite{hoffman1986traveling}. The problem is as follows: given a list of cities and a distance between each pair of cities, find the shortest possible path that visits every city exactly once and returns to the starting city. The TSP has broad applications including: shortest-path
for lasers to sculpt microprocessors and delivery logistics for mail services, to name a few.

The TSP is an area of active research. In fact, several
variants have been derived from the original TSP. In this
paper, we focus on the Euclidean TSP. In the Euclidean TSP, the vertices correspond to points in a $d$-dimensional
space, and the cost function is the Euclidean distance. That is, the Euclidean distance between two cities $x = (x_1, x_2, ..., x_d)
, y = (y_1, y_2, ..., y_d)$ is:
\begin{equation}
	(\sum_{i=0}^{d}(x_i-y_i)^2)^{1/2}
\end{equation}
This simplification allows us to survey several cornerstone algorithms without introducing complex scenarios. The remainder of this paper is organized
as follows. In Section~\ref{sec:background}, we briefly review the first solutions and survey variants to the TSP. We
describe the algorithms used in our experiment in Section~\ref{sec:algo}. A description of
the benchmark datasets and results of the experiment are
detailed in Section~\ref{sec:exp}, and explains the
findings and compares the performance of the algorithms.
We then conclude and describe future work in Section~\ref{sec:conc}.

\section{Background}\label{sec:background}
\begin{figure}
\centering
\includegraphics[width=.7\linewidth]{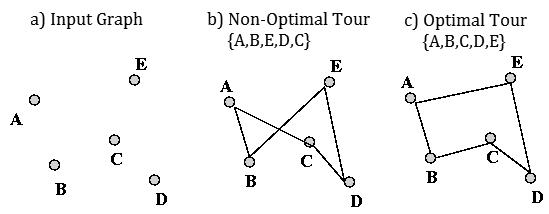}
\caption{}
\label{fig:egtsp}
\end{figure}

An example TSP is illustrated in Figure~\ref{fig:egtsp}. The input is a collection of cities in the two dimensional space. This input can be represented as a distance matrix for each pair of cities or as a list of points
denoting the coordinate of each city. In the latter method,
distances are calculated using Euclidean geometry. A non-optimal tour is shown in sub-figure (b). Although not shown in the figure, each edge will have some non-negative edge
weight denoting the distance between two nodes or cities.
Due to the computational complexity of the TSP, it may be
necessary to approximate the optimal solution. The optimal
tour is shown in sub-figure (c). For small graphs, it may be
possible to perform an exhaustive search to obtain the optimal solution. However, as the number of cities increases, so
does the solutions space, problem complexity, and running
time.

If $n$ is the number of cities. The number of possible edges is $\sum_{i=0}^{n-1}i$. The number of possible tours is $(n - 1)!/2$. since the same tour, with start point
X and Y appears twice: once with X as the start node and
once with Y as the start node.

The TSP was first formulated in the 1930s by Karl Menger in Vienna and Harvard. 
By the mid-1950s, solutions for TSP began to appear. 
The first solution was published by Dantzig, Fulkerson, and Johnson using a dataset of 49 cities. 
In 1972, Richard M. Karp proved that the Hamiltonian cycle problem was NP-Complete, 
which proves that the TSP is NP-Hard.

In modern day, the TSP has a variety of applications to numerous fields. Examples among
these applications include genome sequencing, air traffic control, supplying manufacturing lines, and optimization.

\section{Algorithms}\label{sec:algo}
We now move to a discussion of the algorithms used in our evaluation. 
First, we describe an upper bound for TSP in Section~\ref{sec:rand}.
The traditional greedy and 2-opt approaches are discussed in Section~\ref{sec:greedy} and Section~\ref{sec:2opt}. 
We finally discuss the genetic algorithm in Section~\ref{sec:genetic}.

\subsection{Random Path}\label{sec:rand}
Finding the worst case of TSP is as hard as the best one.
So we uniformly generate a random path for all available edges, 
and use this as a upper bound of optimal path benchmark for all other algorithms.

\subsection{Greedy Algorithm}\label{sec:greedy}
The greedy heuristic is based on Kruskal’s algorithm to
give an approximate solution to the TSP \cite{kim1998comparison}. The algorithm
forms a tour of the shortest route and can be constructed if and only if:
The edges of the tour must not form a cycle unless
the selected number of edges is equal to the number of
vertices in the graph.
The selected edge (before being appended to the tour)
does not increase the degree of any node to be more
than 2.
The algorithm begins by sorting all edges from least weight
to most heavily weighted. After the edges are sorted, the
least heavily-weighted edge is selected and it is added to the tour if it does not violate the above conditions. The algorithm continues by selecting the next least-cost edge and adding it to the tour. This process is repeated until all vertices can be reached by the tour. The result is a minimum
spanning tree and is a solution for the TSP. The runtime for
the greedy algorithm is $O(n^2log(n))$ and generally returns a solution within 15-20\% of the Held-Karp lower bound \cite{rosenkrantz1977analysis}.

\subsection{2opt Algorithm}\label{sec:2opt}
In optimization, 2-opt is a simple local search algorithm first proposed by Croes in 1958 for solving the TSP \cite{croes1958method}.
The main idea behind it is to take a route that crosses over itself and reorder it so that it does not.

A complete 2-opt local search will compare every possible valid combination of the swapping mechanism. This technique can be applied to the travelling salesman problem as well as many related problems. These include the vehicle routing problem (VRP) as well as the capacitated VRP, which require minor modification of the algorithm.

This is the mechanism by which the 2-opt swap manipulates a given route:
\begin{enumerate}
\item take route[1] to route[i-1] and add them in order to new route
\item take route[i] to route[k] and add them in reverse order to new route
\item take route[k+1] to end and add them in order to new route
\item return new route
\end{enumerate}
 
%

\subsection{Genetic Algorithm}\label{sec:genetic}
Genetic algorithms (GA) are search heuristics that at-
tempt to mimic natural selection for many problems in optimization and artificial intelligence \cite{grefenstette1985genetic}. In a genetic algorithm, a population of candidate solutions is evolved over
time towards better solutions. These evolutions generally
occur through mutations, randomization, and recombination. We define a fitness function to differentiate between
better and worse solutions. Solutions, or individuals, with
higher fitness scores are more likely to survive over time.
The final solution is found if the population converges to a solution within some threshold. However, great care must
be taken to avoid being trapped at local optima.

We will now apply a genetic algorithm to the TSP \cite{bryant2000genetic}. We define a fitness function F as the
length of the tour. Supposed we have an ordering of the
cities $A = {x_1 , x_2 , ..., x_n }$ where n is the number of cities.
The fitness score for the TSP becomes the cost of the tour
$d(x, y)$ denote the distance from x to y.
\begin{equation}
	F(A) =\sum_{i=0}^{n-1} d(x_i , x_{i+1}) + d(x_n , x_0 )
\end{equation}
The genetic algorithm begins with an initial, $P_0$ , random
population of candidate solutions. That is, we have a set of
paths that may or may not be good solutions. We then move
forward one time step. During this time step, we perform a
set of probabilistic and statistical methods to select, mutate,
and produce an offspring population, $P_1$ , with traits similar
to those of the best individuals (with the highest fitness)
from $P_0$ . We then repeat this process until our population becomes homogeneous.

The running time of genetic algorithms is variable and dependent on the problem and heuristics used. However, for
each individual in the population, we require $O(n)$ space for
storage of the path. For genetic crossover, the space requirement remains $O(n)$. The best genetic algorithms can find solutions within 2\% of the optimal tour for certain graphs \cite{homaifar1992schema}.

\section{Experiment}\label{sec:exp}
\begin{figure}
\centering
\includegraphics[width=\linewidth]{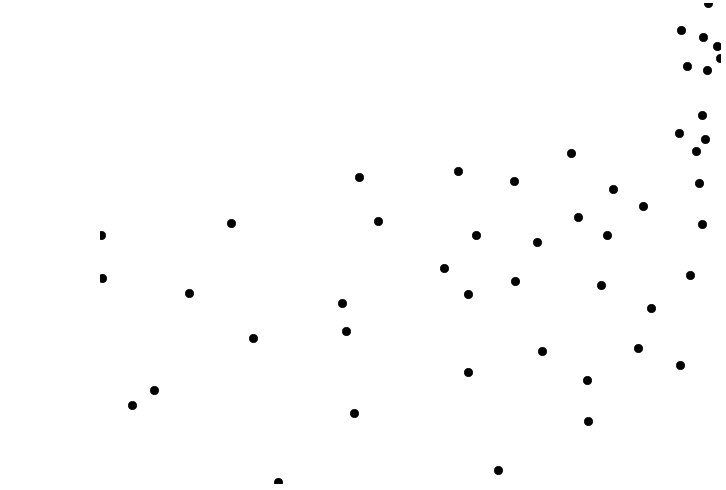}
\caption{the ATT48 dataset in the 2D plane. (the United States)}
\label{fig:00001}
\end{figure}

We benchmark our algorithms using publicly available
datasets. 
Additionally, to test the scalability of the algorithms, we generated a synthetic dataset consisting of 200 cities. 
In all dataset names, the numeric digits represent
the number of cities in the dataset. 
The datasets are as follows: P15, ATT48, and R200.
All datasets except R200 can be found online \cite{data0,data1}. The
ATT48 and SGB128 datasets represent real-data consisting
of locations of cities in the United States. 
A visual representation of the ATT48 dataset in the 2D plane is shown
in Figure~\ref{fig:00001}

Not all datasets have a known optimal tour. When this is the case, we use random path algorithm to infer a upper bound of the optimal tour. 

\subsection{Random Dataset}
The R200 dataset was generated by plotting 200 random, 
uniformly distributed points $(x, y)$, in $R^2$ with $(x, y) \in
[0, 4000]$. As a result, all
distances satisfy the triangle inequality and this dataset can
be classified as a Euclidean TSP dataset. The running time
for creating the dataset is $O(n)$. 
The output is a list of all cities represented as $(x,y)$ points.

\subsection{Comparison}
\begin{figure}
\centering
\includegraphics[width=\linewidth]{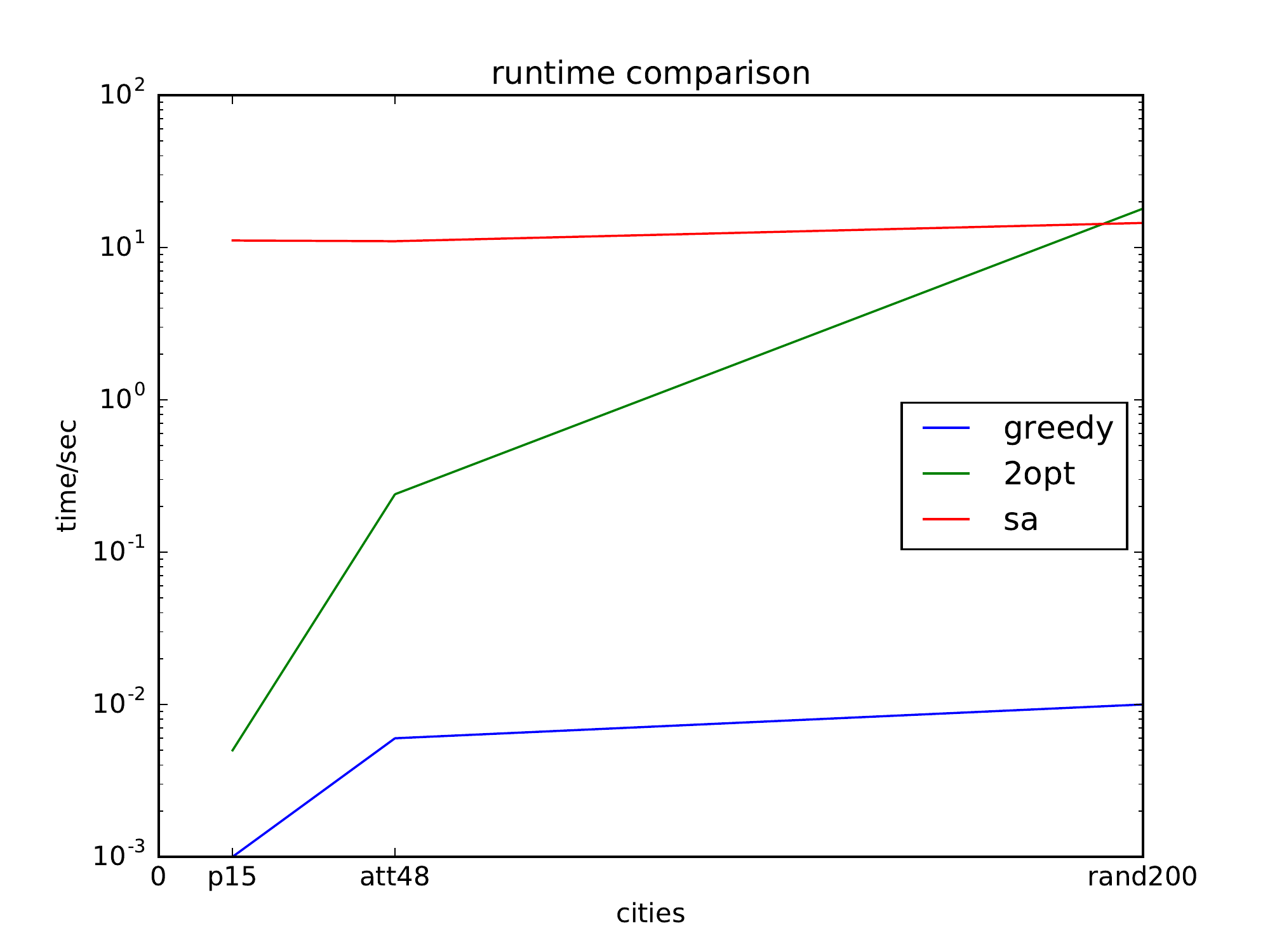}
\caption{runtime comparison, the y axis is in log scale}
\label{fig:tsp}
\end{figure}

\begin{figure}
	\centering
	\includegraphics[width=\linewidth]{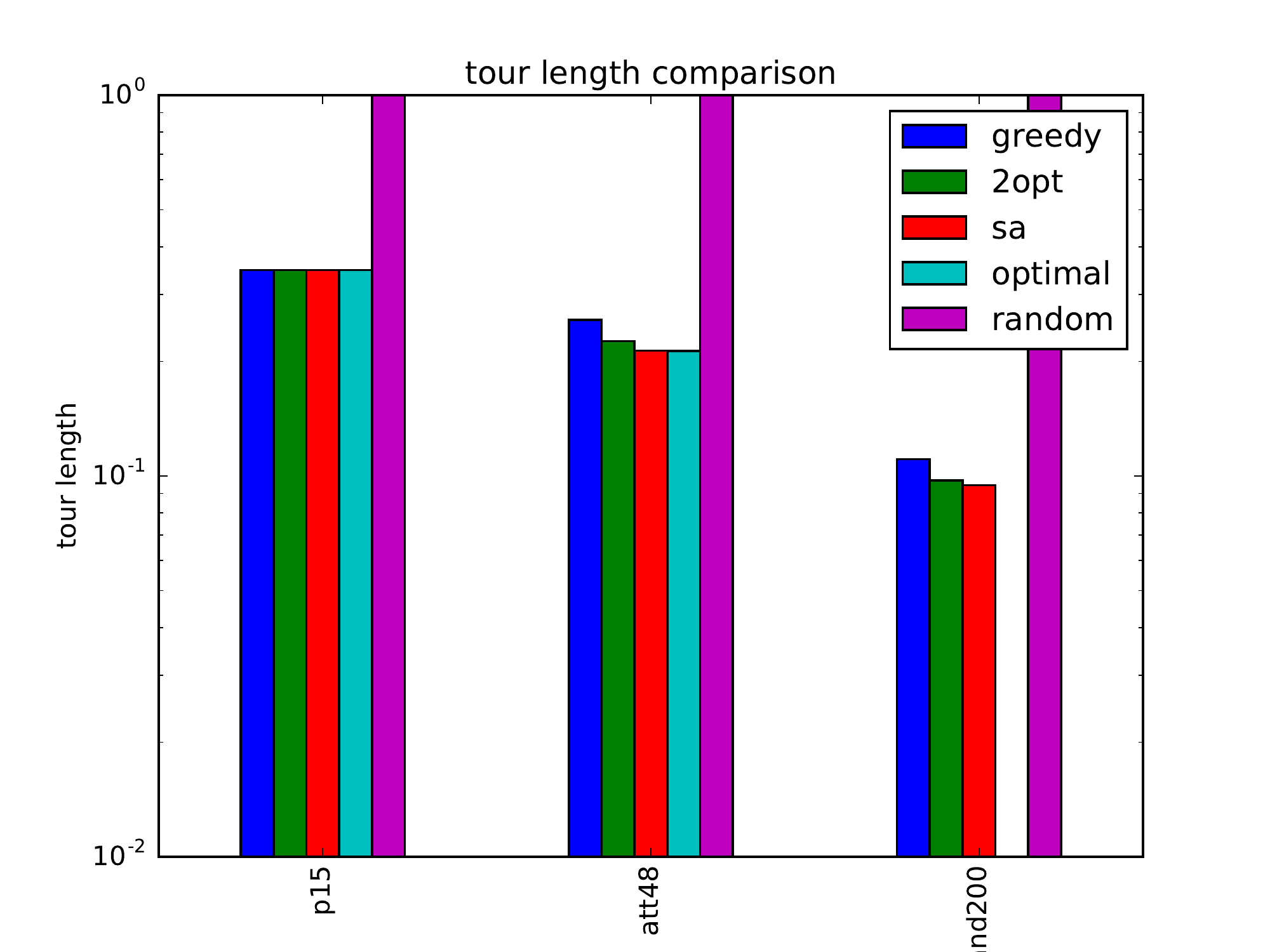}
	\caption{tour length comparison, the y axis is in log scale, divided by random tour length}
	\label{fig:length}
\end{figure}

As we can see in Figure~\ref{fig:tsp}, the greedy is the most efficient. In Figure~\ref{fig:length},
we can see that most algorithms return a solution similar with the optimal for small datasets and become worse for larger datasets.

In terms of running time (Figure~\ref{fig:tsp}), the best algorithm is greedy algorithm. However, in terms of optimal tour length of solution, the best algorithm is GA. This is in line with our expectations and alludes to the fact that different heuristics
are better suited for different situations.

\begin{figure}
\centering
\includegraphics[width=1\linewidth]{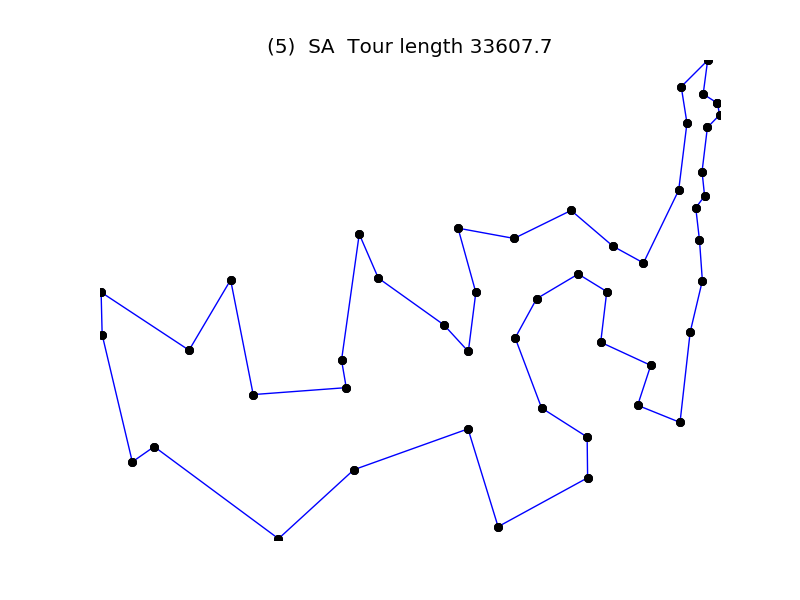}
\caption{Solution generated by genetic algorithm for att48 dataset}
\label{fig:saatt}
\end{figure}

As shown in Figure~\ref{fig:length}, genetic algorithm performs fairly consistently, in comparison to the 2-opt and greedy algorithms, across all datasets. Highlighted in
Figure~\ref{fig:tsp}, the running time of genetic is almost linear. This suggests that for larger
datasets, if running time is a concern, then the genetic
algorithm should be used. Figure~\ref{fig:length} further demonstrates
that genetic algorithm maintains a smaller percent above
optimal than the other algorithms. From this, we can see
that genetic algorithm has high accuracy and better complexity than other heuristics, especially for larger datasets. Surprisingly, genetic algorithm got the optimal solution for att48 dataset, shown in Figure~\ref{fig:saatt}.

\section{Conclusion}\label{sec:conc}
Most of our algorithms attempt to solve the TSP in a
linear fashion. Originating from artificial intelligence, the
genetic algorithm is very different compared to greedy, and 2-opt. Literature suggests that the
best algorithms focus on iteration and convergence to find
optimal tours -- something genetic algorithms attempt to
achieve. For example, the Large Step Markov Chain \cite{hong1997improved}
relies on Markov chains to find convergence of many paths
to form a global optimum and several papers cite Markov
Chains as the best known solution to TSP. Recent studies include using adaptive Markov Chain Monte Carlo algorithms \cite{qiu2008adaptive}. Many of these extend the Metropolis algorithm \cite{homaifar1992schema}, 
a simulated annealing algorithm which attempts to mimic randomness with particles as the temperature varies. This further supports our conclusion that algorithms inspired from
artificial intelligence perform well for finding solutions for the TSP. 
However, these may not be suitable when a guarantee is required.

In this paper, we surveyed several key cornerstone approaches to the TSP. We selected
four well-known algorithms and tested their performance on
a variety of public datasets. Our results suggest that genetic algorithms (and other approaches from artificial intelligence) are able to find a near-optimal solution.

{\small
\bibliographystyle{ieee}
\bibliography{egbib}
}

\end{document}